\documentclass[letterpaper, 10 pt, conference]{ieeeconf}  

\IEEEoverridecommandlockouts                              

\usepackage{graphics} 
\usepackage{epsfig} 
\usepackage{mathptmx} 
\usepackage{times} 
\usepackage{amsmath} 
\usepackage{amssymb}  
\usepackage[normalem]{ulem}
\usepackage{booktabs}
\usepackage{cite}
\usepackage[table]{xcolor}
\usepackage{multirow}  
\useunder{\uline}{\ul}{}
\title{\LARGE \bf
Local Representative Token Guided Merging for Text-to-Image Generation
}
\author{Min-Jeong Lee$^{1}$, Hee-Dong Kim$^{1}$, and Seong-Whan Lee$^{1}$
\thanks{*This research was supported by the Institute of Information \& Communications Technology Planning \& Evaluation (IITP) grant, funded by the Korea government (MSIT) (No. RS-2019-II190079 (Artificial Intelligence Graduate School Program (Korea University)), and No. RS-2024-00457882 (AI Research Hub Project)).}
\thanks{$^{1}$M.-J. Lee, H.-D. Kim, and S.-W. Lee are with the Department of Artificial Intelligence, Korea University, Anam-dong, Seongbuk-ku, Seoul 02841, Korea.
    {\tt\small \{mj\_lee, hd\_kim, sw.lee\}@korea.ac.kr}}
}

\usepackage{graphicx}  
\usepackage{float}
\usepackage{algorithm}
\usepackage{algpseudocode}
\usepackage{amsmath}

\begin{document}

\maketitle
\thispagestyle{empty}
\pagestyle{empty}

\begin{abstract}
Stable diffusion is an outstanding image generation model for text-to-image, but its time-consuming generation process remains a challenge due to the quadratic complexity of attention operations. 
Recent token merging methods improve efficiency by reducing the number of tokens during attention operations, but often overlook the characteristics of attention-based image generation models, limiting their effectiveness.
In this paper, we propose local representative token guided merging (ReToM), a novel token merging strategy applicable to any attention mechanism in image generation. To merge tokens based on various contextual information, ReToM defines local boundaries as windows within attention inputs and adjusts window sizes. Furthermore, we introduce a representative token, which represents the most representative token per window by computing similarity at a specific timestep and selecting the token with the highest average similarity. This approach preserves the most salient local features while minimizing computational overhead. Experimental results show that ReToM achieves a 6.2\% improvement in FID and higher CLIP scores compared to the baseline, while maintaining comparable inference time. We empirically demonstrate that ReToM is effective in balancing visual quality and computational efficiency.

\end{abstract}

\begin{figure*}[t] 
  \centering
  \includegraphics[width=1\textwidth]{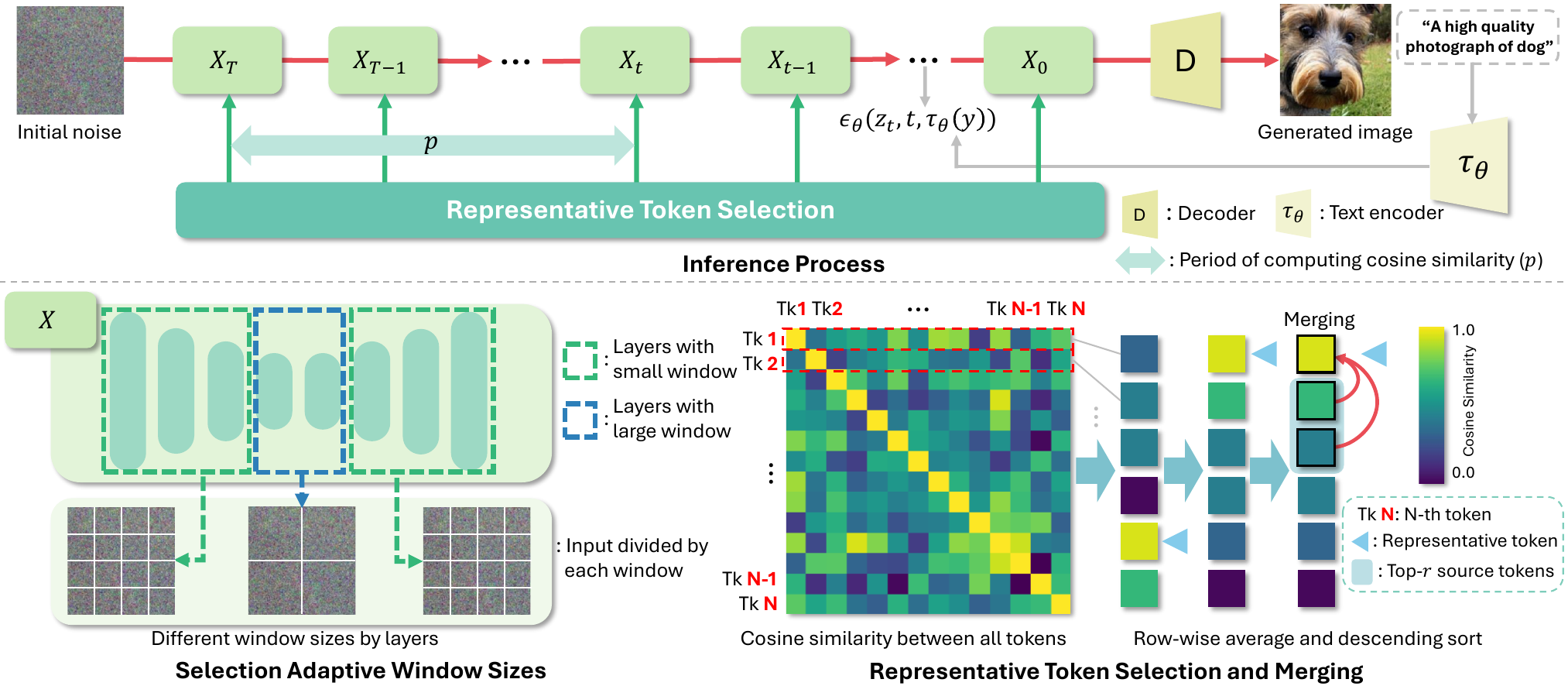}
  \caption{
  Overview of local representative token guided merging. First, we adjust the window size for each transformer block as illustrated in the bottom left diagram, and like the bottom right diagram, compute the cosine similarity between all tokens from TK \(1\) to TK \(N\). Then, we average the similarity values and select representative tokens. Among the remaining tokens except for the representative token, the top-\(r\) source tokens are merged into the representative token, and \(r\) is determined by the set merge ratio. This process is periodically applied to U-Net with a predefined period \(p\) during the inference process.}
  \label{fig2} 
\end{figure*}
\section{INTRODUCTION}
Recent advances in text-to-image generation have been greatly facilitated by LDMs~\cite{rombach2022high}, and stable diffusion has emerged as a state-of-the-art approach.
These models leverage a latent space to synthesize high-quality images while being more computationally efficient than pixel-space generative approaches~\cite{goodfellow2020generative, ho2020denoising, song2020denoising}. Despite these advantages, generative models still suffer from high memory consumption and slow inference, primarily due to the high computational cost of token processing and redundant operations in the attention mechanism. This presents a major challenge for deploying these models in real-world applications where efficiency is critical~\cite{kim2015abstract, lee1990translation, roh2007accurate}. One of the main computational bottlenecks in stable diffusion lies in its U-Net~\cite{ronneberger2015u} with transformer-based blocks, which employ self-attention mechanisms to capture global dependencies~\cite{vaswani2017attention}. This is because self-attention requires calculations for every pair of tokens, resulting in quadratic complexity as the number of tokens increases. This issue becomes particularly pronounced when generating high-resolution images, leading to active research on improving the computational efficiency of diffusion models while preserving generative performance. To address this issue, various approaches such as knowledge distillation~\cite{meng2023distillation,salimans2022progressive}, token pruning~\cite{ yin2022vit,kim2022learned, ryoo2021tokenlearner} and token reduction~\cite{dou2023tore} have been explored. 
However, most existing studies improve efficiency in ways that require internal modifications to the model structure or subsequent fine-tuning or retraining~\cite{chen2021trainonceoneshotneural, lim2000text,fang2023structuralpruningdiffusionmodels}. Among these, token merging mechanisms~\cite{bolya2022token} have been proposed as an effective technique to reduce computational cost while maintaining generative performance without any additional training and tuning. Token merging, initially applied only to classification tasks in ViT~\cite{dosovitskiy2020image}, has claimed that it could also be effective for downstream tasks such as generative models. Based on this assumption,~\cite{bolya2023token} applied the method to diffusion models leveraging token merging in attention operations by grouping tokens with similar features, thereby reducing the number of tokens processed. While this method has demonstrated speedup and a reduction in memory usage, it was not specifically tailored to the structural and operational characteristics of generative models with attention mechanism, potentially missing opportunities to further enhance generation quality and computational efficiency. Notably, this method applies a fixed merging strategy across all layers, and does not account for the characteristic of U-Net in diffusion models~\cite{yang2021focal,luo2016understanding}.
Additionally, merging are based solely on randomly selected destination tokens and pairwise similarity, which may overlook the structural coherence of the image representations.

To mitigate these existing shortcomings, we propose ReToM, a novel token merging strategy that preserves local representations while generating high-quality images. 
Unlike existing approaches, our method defines a local boundary as a window within attention input and adjusts the window size in diverse ways. Instead of randomly selecting destination tokens, we introduce a selection mechanism that identifies a representative token based on its ability to best retain the local feature information within each merging window. 
To verify the effectiveness of the proposed representative token selection and adaptive window size adjustment, we conducted experiments comparing our method with existing token merging methods. Our extensive experimental evaluations demonstrate that our method reduces redundant operations while maintaining the fine-grained structure and global consistency of generated images, providing higher quality results compared to existing methods. Compared to the previous random selection, the model that applied the representative token selection strategy within the varying window size recorded better performance in FID. Although the similarity calculation for additionally quadratic complexity is required, it contributes to the improvement in generation performance without increasing the inference time by applying our similarity computation caching strategy considering the iterative process of the diffusion model. These experimental results empirically verify that the representative token selection strategy effectively improves token merging performance in image generative models.
By introducing adaptive window size with local boundary and representative token selection, our approach enhances the quality and efficiency of generative models while optimizing computational performance. It is particularly well-suited for applications such as real-time image synthesis, high-resolution text-to-image generation, and model compression in resource-constrained environments.

The contributions of this paper are summarized as follows: 
\begin{itemize} 
    \item We demonstrate that similarity-based token selection enables merging with preservation of essential local representations within each window, thereby improving image generation performance.
    \item We propose a novel token merging strategy tailored for attention-based models, where local boundaries within attention inputs are defined, enabling merging operations with various contextual information through the adapting window size.
    \item Our method significantly improves performance compared to existing token merging techniques without increasing computational cost.
\end{itemize}

\section{RELATED WORK}

\subsection{Diffusion Model}
Diffusion models have demonstrated exceptional performance in the field of image generation. DDPM~\cite{ho2020denoising} was designed to generate images through an iterative denoising process. To overcome the high computational costs, DDIM~\cite{song2020denoising} improved sampling efficiency, and LDM performed denoising in latent space to reduce computational overhead. These diffusion models still face high computational complexity due to repetitive sampling and quadratic attention operations~\cite{vaswani2017attention, rombach2022high}.
Prior works have mainly focused on reducing denoising steps~\cite{li2023autodiffusiontrainingfreeoptimizationtime, lu2022dpm} or optimizing attention computation within transformer blocks~\cite{ramachandran2019stand}. Nevertheless, the quadratic attention operation still causes large computational costs~\cite{rabe2021self}, and optimization studies for this issue are relatively insufficient. Among them, various studies have explored token pruning~\cite{kim2022learned, ryoo2021tokenlearner}, while others have focused on token merging~\cite{bolya2022token}. 
Our work follows the token merging of research with the aim of reducing attention operations.

\subsection{Token Merging}

Token merging has been explored in transformers as an effective way to reduce computational complexity while maintaining competitive performance~\cite{Liang2022NotAP, yin2022vit}. Previous studies have demonstrated that merging redundant tokens during inference significantly improves the efficiency of ViT, maintaining performance~\cite{bolya2022token, xu2022groupvit}.
Building on these findings, token merging has been applied to the attention operation of diffusion models, particularly stable diffusion~\cite{bolya2023token}.
Although this approach has shown reduced computational overhead, it does not fully account for the structural properties of attention specific to diffusion models~\cite{bolya2022token}.
Importance-based token merging~\cite{wu2024importance} utilizes classifier-free guidance information, but does not account for spatial locality between tokens and is unstable in early timesteps. ATC~\cite{haurum2024agglomerative} employs a parameter-free hierarchical clustering-based method. In attention-based downstream tasks, optimal strategies for token merging remain an open challenge~\cite{lee1995multilayer}, requiring further improvements to effectively balance computational efficiency and generative quality.
\begin{algorithm}[!t]
\caption{Representative Token Selection and Merging}
\begin{algorithmic}[1]
\State \textbf{Input:} Merging ratio \( R \), timestep \( t \), period \( p \), merging weight \(\alpha\), total windows \(\mathcal{W}\)
\State \textbf{Init:} \( t = 0 \), cached similarity \( {sim}_{\text{cache}} \), the number of source tokens \(r\)
\State Divide input tokens into window \( W \in \mathcal{W} \)
\For{window \( W \in \mathcal{W} \)}

    \(r \leftarrow \text{the number of tokens in \(W\)}\cdot R\)
    \If{$t \bmod p = 0$}
        \State Compute cosine similarity:
        \[
        {cosSim}(x_i,x_j) = \frac{x_i \cdot x_j}{\|x_i\| \cdot \|x_j\|}
        \]
        \State Compute average similarity:
        \[
        ~~~~~~~~~~{sim}(x_i,W) = \frac{1}{N_W-1} \sum_{j \in W, j \ne i} {cosSim}(x_i,x_j)~~~~~~~~~~\eqref{eqn:algo}
        \]
        \State Cache similarity: \( {sim}_{\text{cache}} \leftarrow {sim}(x_i,W) \)
    \Else
        \State Use cached similarity: \( {sim}(x_i,W) \leftarrow {sim}_{\text{cache}} \)
    \EndIf

    \State \textbf{Select Tokens:}
    \State \( D \leftarrow {sim}(x_i,W)[0] \) \Comment{Most representative token}
    \State \( S \leftarrow {sim}(x_i,W)[1:r] \) \Comment{Top-\( r \) similar tokens}

    \State \textbf{Merge Tokens:}
    \vspace{-4mm}
    \[
    ~~~~~~~~~~{x}_{\text{merged}} = \alpha D + (1 - \alpha) \cdot \frac{1}{r} \sum S_i~~~~~~~~~~~~~~~~\eqref{merge}
    \]
    \vspace{-6mm}   
    \State \( t \leftarrow t + 1 \)
\EndFor    
\end{algorithmic}
\label{alg1}
\end{algorithm}
\section{METHOD}

ReToM is designed to adapt token merging to diffusion models while preserving the expressiveness of essential information. Fig.~\ref{fig2} provides an overview of our token merging method. Unlike the previous merging approach with fixed-size regions~\cite{bolya2023token}, we introduce a window as a local boundary and adjust their size variously. ReToM also selects a representative token based on local similarity. Additionally, our approach incorporates a similarity computation caching strategy to balance computational efficiency and effectiveness in token merging. The following sections provide a detailed explanation of each component of our proposed method, including adaptive window size, local token merging with the representative token, and a similarity computation caching strategy.
\begin{table*}[t]
\centering
\caption{Evaluation of ReToM with different token merging strategies applied to Stable Diffusion. The table presents the FID, CLIP score, and generation speed (s/im) for different strategies including variations in window size and Dest. Token selection. }

{\scriptsize
\renewcommand{\arraystretch}{1.2}
\setlength{\tabcolsep}{14pt}
\resizebox{\textwidth}{!}{%
\begin{tabular}{lccccc}
\toprule
 & \multicolumn{2}{c}{\textbf{Methods}} & & & \\ 
\cmidrule(lr){2-3}
\textbf{Models} & \textbf{Window Size} & \textbf{Dest. Token} & \textbf{FID (↓)} & \textbf{CLIP (↑)} & \textbf{s/im} \\
\midrule
\midrule
Stable Diffusion~\cite{rombach2022high} & – & – & 37.02 & 38.10 & 2.62 \\ 
\midrule
ToMeSD~\cite{bolya2023token} & Fixed (2) & Random & 37.20 & 36.00 & \textbf{2.10} \\ 
\midrule
ATC~\cite{haurum2024agglomerative} & – & – & 36.96 & 37.47 & 110.25 \\ 
\midrule
\multirow{3}{*}{ReToM w/ Fixed Window} 
& Fixed (2) & Rep. Token & \underline{35.00} & 37.90 & 2.25 \\
& Fixed (8) & Rep. Token & 35.20 & 37.05 & 2.19 \\
& Fixed (16) & Rep. Token & 36.53 & 37.73 & 2.17 \\
\midrule
ReToM w/ Adaptive Window & Adaptive & Random & 37.11 & \underline{38.81} & \underline{2.15} \\
\midrule
ReToM (Ours) & Adaptive & Rep. Token & \textbf{34.89} & \textbf{39.40} & 2.17 \\
\bottomrule
\end{tabular}
}
\label{table2}
}
\end{table*}

\subsection{Adaptive Window Size with Local Boundary} The existing token merging method fixes the region size for merging across all layers without accounting for layer-wise features.
To address this, we propose a token merging strategy that adaptively adjusts the merging window size, drawing inspiration from the concept of receptive fields in different layers of the U-Net. We define a window as a local boundary of the tokens used to compute the similarity, and one window as a boundary corresponds to one representative token. 
As illustrated in Fig.~\ref{fig2}, the section on selection adaptive window sizes explains that the green dotted regions correspond to the downsampling and upsampling blocks of the U-Net, while the blue dotted region denotes the bottleneck layers. Following the structure of the U-Net, we assign small window sizes to the green regions, where local detail preservation is crucial, and larger window sizes to the blue bottleneck region, where capturing global context is more important.
By setting the adaptive window size, our strategy maintains a balance between local feature retention and global context.

\subsection{Local Token Merging with Representative Token} \label{localtoken}
The random destination token selection and bipartite matching cannot select the most informative feature token in merging, potentially leading to information loss and merging of dissimilar tokens.
To address this issue, we propose a destination token selection strategy that chooses the most representative token in each window based on local feature similarity. As shown in Fig.~\ref{fig2}, representative token selection and merging section describes the computation of pairwise cosine similarity among all tokens within a window. Then, we compute the row-wise average of the similarity matrix, the average cosine similarity for a token \( x_i \) in a window \( W \) is defined as:
\begin{equation}
sim(x_i,W) = \frac{1}{N_W - 1} \sum_{j \in W, j \neq i} {cosSim}(x_i,x_j),
\label{eqn:algo}
\end{equation}
where \( N_W\) is the number of tokens in \( W\). This cosine similarity among all tokens is calculated every period \(p\). The average similarity values are then sorted in descending order, and the token with the highest \( sim(x_i,W) \) is selected as the representative token. 
Once the representative token is selected, according to the predefined merging ratio \(r\), a subset of tokens in the window are selected as source tokens and merged into the representative token.
The merging is performed by computing the mean of the representative token and the source tokens, defined as follows:
\begin{equation}
    {x}_{\text{merged}} = \alpha {D} + (1 - \alpha) \cdot \frac{1}{r} \sum_{i=1}^{r} {S}_i,
\label{merge}
\end{equation}
where \(D\) is the representative token, \(S_i\) are the source tokens and $\alpha$ controls the balance between the representative token and the source tokens. Our method enables the selected token to retain the most representative features of the window, improving information expressiveness and merging stability. By eliminating redundant tokens while preserving essential information within the window, the proposed method improves efficiency and minimizes information loss.

\subsection{Similarity Computation Caching Strategy}
In the stable diffusion model, the input data at each timestep is updated based on the previous timestep, and changes in the latent representation across consecutive timesteps are gradual. This means that that the relative similarity between tokens remains largely consistent. To select representative tokens in Sec. \ref{localtoken}, computing cosine similarity at every timestep leads to unnecessary computational overhead and degrades overall model efficiency.
To address this, we design a similarity computation caching strategy that leverages the gradual change in token similarity across consecutive timesteps. As illustrated in the inference process in Fig.~\ref{fig2}, the cosine similarity of all tokens within a window is computed and cached with a period of \(p\), and reused across subsequent timesteps for efficiency. We show an analysis that the similarity of representative tokens between successive timesteps through experiment on real-world data in Sec. \ref{exper_caching}. The detailed algorithm for this process is provided in Algorithm~\ref{alg1}.

\section{EXPERIMENTS}
In this section, we evaluate the performance of ReToM in text-to-image generation, aiming to reduce the computational burden while preserving the quality of the image. We apply ReToM to the transformer blocks of the U-Net in the stable diffusion model. Note that ReToM does not require additional model training or fine-tuning, allowing direct integration into existing attention-based models.

\subsection{Experimental Setting}

All experiments were conducted under the same conditions using stable diffusion model. The number of sampling steps was set to 50, and the guidance scale was fixed at 7.5. Images were generated at a resolution of 512×512. 
Our merging strategy was applied across all transformer block layers, where representative tokens were selected based on token similarity metrics. ImageNet~\cite{deng2009imagenet} validation dataset, which consists of 1,000 classes with a total of 50,000 images, was used for the experiments. 

\subsection{Evaluation Metrics}
To evaluate our method, we compare the generated images with real ones using samples from the ImageNet validation dataset. We select five real images per class, resulting in a total of 5,000 images, and generate 2,000 images, two images per class using ReToM. We measure the FID score to evaluate the quality of the generated images by computing the statistical distance between the real and generated feature distributions, where a lower FID score indicates better quality. Additionally, we measure the generation time per sample (s/im) to assess computational efficiency. To evaluate semantic alignment between generated images and their corresponding text prompts, we compute the CLIP score~\cite{Hessel2021CLIPScoreAR}, where a higher CLIP score indicates better text-image consistency.

\begin{figure*}[t] 
  \centering
  \includegraphics[width=1\textwidth]{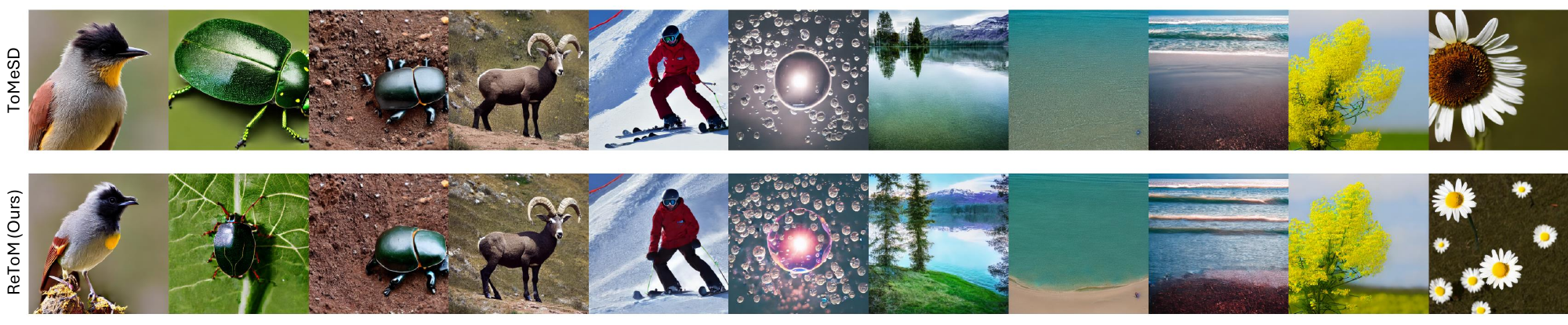}
  \caption{Qualitative results of ToMeSD and our ReToM applied to the stable diffusion model using the ImageNet validation dataset. While ToMeSD focuses on only objects or loses background representation rather than captures both backgrounds and objects, our model efficiently leverages various contextual information to naturally captures backgrounds and objects, maintains good detail without imbalance, and shows overall image consistency.
  }
  \label{fig3} 
\end{figure*}

\begin{table}[]
\centering
\caption{Evaluation of ReToM with the most representative and least representative tokens as destination tokens. The table presents the FID, SSIM score for each token selection.
}
\renewcommand{\arraystretch}{1.5} 
\resizebox{\columnwidth}{!}{%
\begin{tabular}{lccc}
\hline

{\textbf{Model}}                  & {\textbf{Methods}}                           & {\textbf{FID (↓)}}   & {\textbf{SSIM (↑)}}   \\ \hline
\multirow{2}{*}{ReToM} & w/ The Most Rep. Token  & \textbf{34.89} & \textbf{0.74} \\
                       & w/ The Least Rep. Token & 36.26 & 0.72 \\ \hline
\end{tabular}%
}
\label{table3}
\end{table}

\subsection{Adaptive Window Size with Local Boundary}
To demonstrate whether the introduction of localized merging windows enhances token merging efficiency, ReToM defines a local boundary within the attention input as a window and adjusts the window to various sizes. As shown in TABLE~\ref{table2}, ReToM w/ Fixed Window demonstrates robust performance across various window sizes, achieving the best FID (35.00) and CLIP score (37.90), and outperforming the token merging for stable diffusion (ToMeSD)~\cite{bolya2023token} when using a fixed window size of 2. These results indicate that utilizing diverse window sizes in ReToM offers distinct advantages over the fixed-size approach of ToMeSD, as it leverages varying local context to better preserve features and improve image quality. Furthermore, by adopting an adaptive window size strategy, ReToM achieves the best FID (34.89) and CLIP score (39.40) among all configurations, demonstrating that adaptive window mechanisms can effectively balance feature preservation with computational efficiency in diffusion models. 

\subsection{Local Token Merging with Representative Token}
To evaluate that using representative token as destination token in the merging process preserves meaningful content in generated images, we compare it with random selection and evaluate the CLIP score in addition to FID. As shown in TABLE~\ref{table2}, ReToM demonstrates that selecting a  representative token (Rep. Token) leads to performance improvements. Compared to ToMeSD, only selecting destination token (Dest. Token) as a Rep. Token with applying all fixed window size in ReTom improves the FID from 37.20 to a maximum of 35.00 and the CLIP score from 36.00 to 37.90. These results demonstrate that random selection of Dest. Token can result in unnecessary information loss based on cosine similarity, while merging based on the most representative tokens helps retain key structural details, leading to better image quality and improved text-image alignment. This finding is demonstrated in TABLE~\ref{table3}, further validating the importance of Rep. Token selection. When merging is performed into the most Rep. Token in each window, the model achieves best FID and SSIM scores. In contrast, merging into the least Rep. Token leads to degradation in both metrics. These results suggest that selecting the most representative token is crucial for preserving structural consistency and salient features during merging.

\subsection{Similarity Computation Caching Strategy} \label{exper_caching}
\begin{figure}[t!]
\centering
\scriptsize
\centerline{\includegraphics[width=0.5\textwidth]{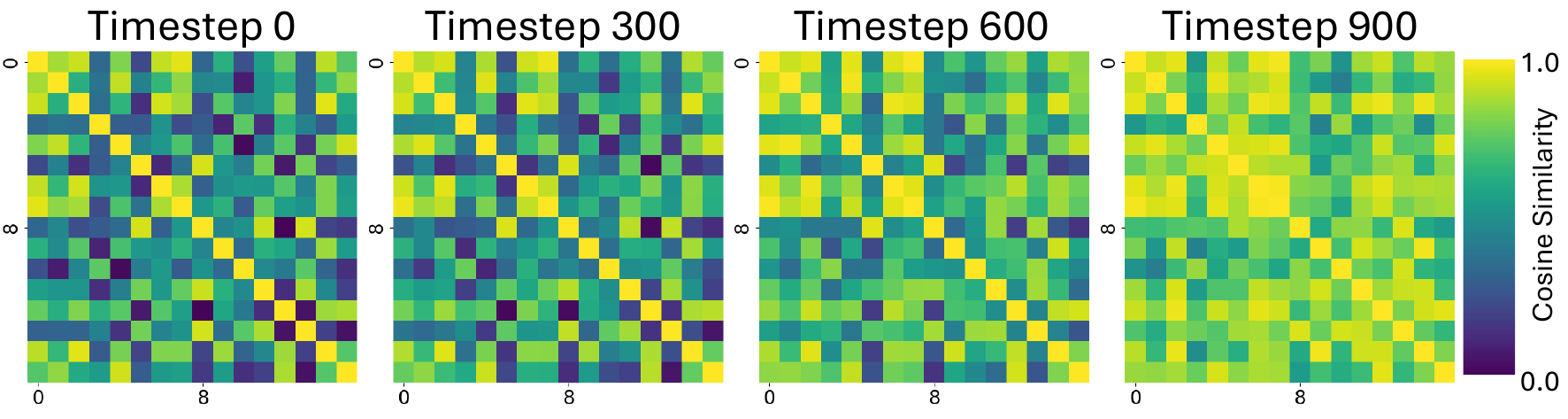}}
\caption{An visualization of changes in the cosine similarity of all tokens within a window as the timestep progresses. From left to right, the overall similarity pattern gradually changes, demonstrating that the cosine similarity between tokens remains consistent across consecutive timesteps.}
\label{fig4} 
\end{figure}

As shown in Fig.~\ref{fig4}, to demonstrate the strategy of calculating and caching cosine similarity at regular timestep period and using it in consecutive timesteps, we visualize heatmaps where each cell represents the cosine similarity between a pair of tokens within the same window at each timestep (0, 300, 600, and 900). We analyze that while the overall similarity pattern gradually evolves over timesteps, the structural consistency between neighboring timesteps is largely preserved. This analysis demonstrates that the similarity between consecutive timesteps remains high, indicating that it is unnecessary to compute the cosine similarity at every timestep. Our approach efficiently leverages this gradual evolution and consistency of diffusion models, improving computational efficiency.

\subsection{Overall Effectiveness of ReToM}
To evaluate the effectiveness of ReToM, we compare its performance with the baseline stable diffusion model, ToMeSD, and ATC. TABLE~\ref{table2} presents ReToM significantly improves image generation quality while maintaining computational efficiency. 
Compared to stable diffusion without token merging, ReToM achieves a 5.8\% improvement in FID (from 37.02 to 34.89) while reducing the computation time from 2.62 s/im to 2.17 s/im. Although ToMeSD achieves faster inference (2.10 s/im), its FID (37.20) is worse than the baseline, indicating a trade-off between efficiency and quality. ATC shows a better FID (36.96) than ToMeSD, but suffers from significantly slower inference when evaluated using the publicly official code, which limits its practicality in real-time generation tasks. In contrast, ReToM outperforms ToMeSD and ATC in FID (34.89) while maintaining a comparable inference speed, thereby achieving a superior balance between computational efficiency and visual fidelity. As shown in Fig.~\ref{fig3}, the ReToM image naturally captures both objects and the background like leaves, grass, soil and sand, and the details tend to be maintained well without shape distortion or imbalance. On the other hand, for images generated by ToMeSD that use a fixed window size without a representative token, there are cases in which the shape of objects is distorted or the background is expressed unnaturally at a specific location.
The consistency of the overall image increases as ReToM more flexibly captures the context by reflecting the characteristics of local boundaries based on representative tokens. In particular, even in images containing complex backgrounds, there is no unnatural conversion between objects and backgrounds.
ReToM shows improvements that effectively preserve key structural information while eliminating redundant computations, enabling it to be a practical solution for real-world text-to-image generation tasks.

\section{CONCLUSION}
In this paper, we proposed ReToM, a novel token merging framework designed to optimize attention-based models with improving both image generation quality and computational efficiency. Our method introduces a similarity-based token selection strategy, enabling merging operations to retain essential local representations within each window. We adjust the window size like varying receptive fields across different U-Net layers. Through extensive experiments using stable diffusion on the ImageNet dataset, we demonstrated that ReToM significantly improves image quality, as evidenced by enhanced FID, CLIP scores. Importantly, our method achieves these improvements without increasing computational complexity, making it a practical solution for real-world applications. Moreover, ReToM does not require additional model training or fine-tuning, allowing direct integration into existing models that utilize attention mechanisms. Lastly, our similarity computation caching strategy reduces redundant computations further contributing to efficiency gains. These results highlight ReToM’s effectiveness in balancing performance and efficiency in diffusion models.
Moving forward, this approach can be extended to other attention-based architectures, and future research may also explore alternative similarity metrics or dynamic merging strategies to further refine token merging efficiency.

\addtolength{\textheight}{-12cm}   



\bibliographystyle{IEEEtran}  
\bibliography{IEEEabrv, ref}
\end{document}